\documentclass[letterpaper, 10 pt, conference]{ieeeconf} 
\IEEEoverridecommandlockouts
\usepackage{mathrsfs}
\usepackage{xcolor}

\usepackage{graphics} 
\usepackage{epsfig} 
\usepackage{amsmath} 
\usepackage{amssymb} 
\usepackage{color}
\usepackage{bm}
\usepackage{subcaption}
\usepackage{multirow}
\usepackage{array}
\usepackage{booktabs}
\usepackage{algorithm,algcompatible}
\usepackage[noend]{algpseudocode}
\usepackage{mathtools}
\usepackage{cite}
\usepackage{makecell}
\usepackage{hyperref}
\usepackage{cuted}
\usepackage{balance}
\definecolor{pointcolor}{RGB}{0,114,189}
\definecolor{linecolor}{RGB}{217,83,25}
\definecolor{planecolor}{RGB}{237,177,32}
\definecolor{spherecolor}{RGB}{119,172,48}
\definecolor{ellipsoidcolor}{RGB}{126,47,142}
\definecolor{cylindercolor}{RGB}{77,190,238}
\definecolor{conecolor}{RGB}{162,20,47}

\graphicspath{{figures/}}

\title{\huge \bf EVLoc: Event-based Visual Localization in LiDAR Maps \\ via Event-Depth Registration}

\author{Kuangyi Chen\quad  Jun Zhang\quad  Friedrich Fraundorfer
\thanks{Kuangyi Chen, Jun Zhang and Friedrich Fraundorfer are with the Institute of Visual Computing, Graz University of Technology, 8010 Graz, Austria. E-mail: \tt\small\{kuangyi.chen, jun.zhang, friedrich.fraundorfer\}@tugraz.at}
}

\begin{document}
\maketitle
\begin{abstract}
Event cameras are bio-inspired sensors with some notable features, including high dynamic range and low latency, which makes them exceptionally suitable for perception in challenging scenarios such as high-speed motion and extreme lighting conditions. In this paper, we explore their potential for localization within pre-existing LiDAR maps, a critical task for applications that require precise navigation and mobile manipulation. Our framework follows a paradigm based on the refinement of an initial pose. Specifically, we first project LiDAR points into 2D space based on a rough initial pose to obtain depth maps, and then employ an optical flow estimation network to align events with LiDAR points in 2D space, followed by camera pose estimation using a PnP solver. To enhance geometric consistency between these two inherently different modalities, we develop a novel frame-based event representation that improves structural clarity. Additionally, given the varying degrees of bias observed in the ground truth poses, we design a module that predicts an auxiliary variable as a regularization term to mitigate the impact of this bias on network convergence. Experimental results on several public datasets demonstrate the effectiveness of our proposed method. To facilitate future research, both the code and the pre-trained models are made available online\footnote{https://github.com/EasonChen99/EVLoc}.


\end{abstract}
\section{Introduction}
Accurate localization techniques are essential for autonomous robots, such as self-driving vehicles and drones. Currently, GPS is widely used for global localization, but its accuracy and stability are compromised when there is no direct line of sight to satellites, such as indoor scenarios. As a result, many researchers are turning to sensors for active localization, which involves determining the 6-Degree-of-Freedom (DoF) poses of robots by comparing online sensor measurements with reference maps, such as 3D point clouds. Visual localization, a subset of this task, uses cameras like monocular or binocular cameras for localization. It is popular due to the compact size and affordability of cameras. While current methods have achieved remarkable performance, they struggle with challenges like motion blur and extreme lighting due to the limitations of conventional cameras. 

Event cameras can overcome these challenges because they have several significant advantages, including an extremely high temporal resolution and low latency, both in the order of microseconds, as well as a remarkable dynamic range (140 dB compared to the 60 dB typical of conventional cameras), and low power consumption. Given these features, event cameras hold a  considerable potential for use in scenarios that pose challenges for conventional cameras\cite{gallego2020event}. Event-based visual localization techniques have emerged over the past few years. Some of them reconstruct intensity\cite{fischer2020event} or edges\cite{9635907, yuan2016fast} from events before feature matching, while others manually construct\cite{fischer2022many} or use learning-based descriptors\cite{kong2022event}. Additionally, some methods utilize neural networks to directly regress event camera poses\cite{nguyen2019real, jin20216, lin20226, ren2024simple}, usually inspired by frame-based camera pose estimation approaches like DSAC\cite{brachmann2017dsac}. However, in the above approaches, the reference maps for localization are either based on a pre-established database or implicitly encoded in the model parameters. The former requires much labor for each new scene, while the latter struggles with very limited generalization.


\begin{figure*}[ht]
    \centering
    \includegraphics[width=0.9\linewidth]{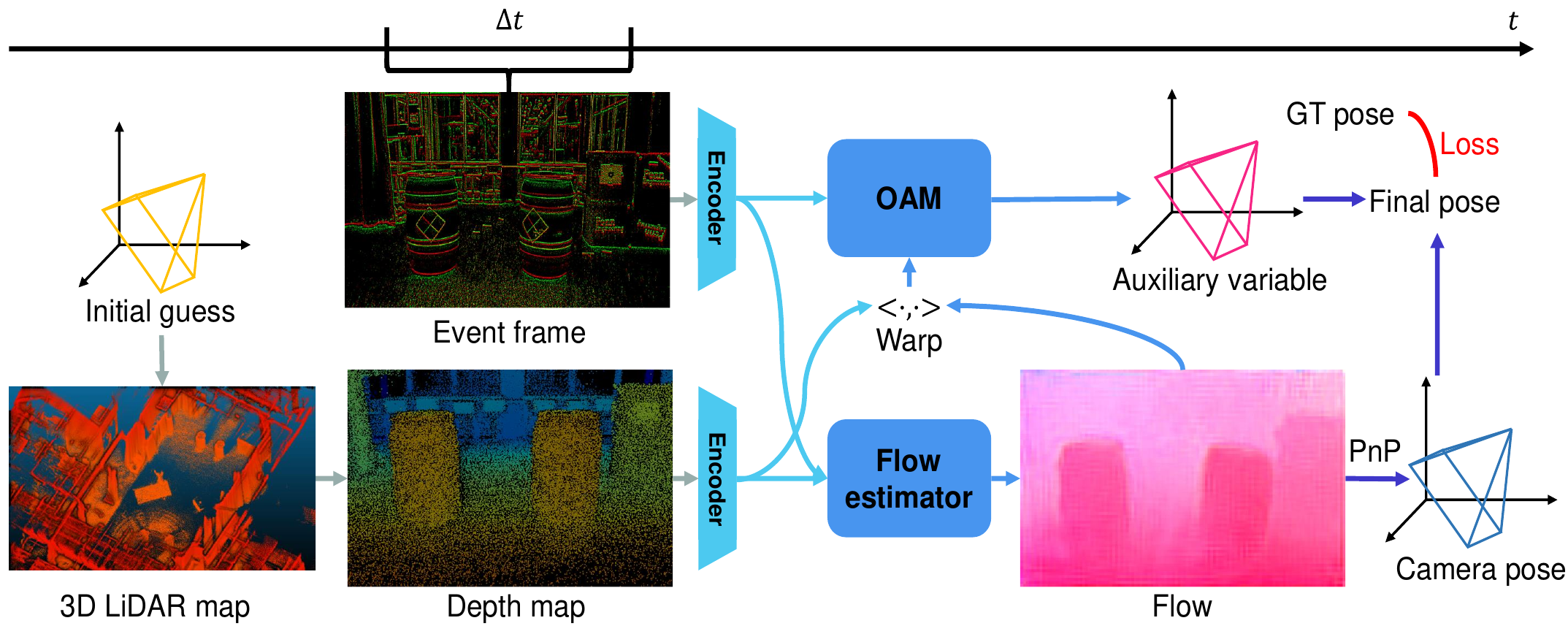}
    \caption{Overview of the proposed EVLoc. We assume that a coarse initial pose guess is available to serve as the starting point for precise localization. We project the 3D LiDAR map into 2D space to generate a depth map based on this initial pose. Simultaneously, events within a fixed time interval $\Delta t$ are converted into an event frame. The event frame and depth map are then input into the flow estimator to obtain the event-depth flow, which is used to warp the encoded depth features. These warped depth features, along with the encoded event features, are fed into the Offset Alleviation Module (OAM) to predict the auxiliary variable. Finally, a PnP solver calculates the camera pose from the 2D-3D correspondences obtained from the estimated flow.} 
    \label{fig:framework}
    \vspace{-5mm}
\end{figure*}

In this paper, we propose an event-based visual localization framework that estimates 6-DoF poses by establishing 2D-3D correspondences between events and an existing LiDAR map. We leverage off-the-shelf LiDAR maps as reference maps, which are easily available nowadays. The primary challenge lies in the inherent modality gap between events and LiDAR points. Inspired by recent frame-based approaches for camera localization in LiDAR maps\cite{Cattaneo2020CMRNetMA, CHEN2022209, yu2024i2d}, we propose to employ existing advanced optical flow estimation networks for event-depth registration. Specifically, a rough initial pose estimate is required in our framework, which can be obtained through coarse global localization techniques such as visual place recognition approaches, wheel odometry, etc. We first project the LiDAR points into 2D space to generate a depth map based on the initial pose, while transforming the event stream into a frame-based representation. Time surface is commonly used for event-based image processing\cite{benosman2012asynchronous}, which aggregates events accumulated over regularly spaced intervals into a single frame. However, due to the arbitrariness of motion and the noisy nature of event cameras, the resulting event frames often contain blurred scene structures. Since events are typically triggered at high-gradient pixels (edges) and LiDAR depth captures the geometrical structure of the scene, we argue that clear structural information is crucial for effective event-depth registration. 
To enhance structural details and suppress background noise, we develop a novel event representation derived from the typical surface representation.
Then, we utilize an RAFT-based optical flow estimation network\cite{Teed2020RAFTRA} to estimate correspondences between the event frames and the depth maps. During training, we observe varying degrees of bias existing in the given ground truth poses, leading to offsets (misalignments) between the event frames and the projected depth maps. To address this issue, we design a module that predicts an auxiliary variable as a regularization term to mitigate this bias during training. Finally, we utilize a PnP solver to calculate the camera poses based on the estimated event-depth correspondences.

The main contributions are listed as follows:

\begin{itemize}

\item We introduce the first framework for event-based visual localization in LiDAR maps, enabling localization through event-depth registration using an optical flow estimation network.

\item We develop a novel frame-based event representation that enhances structural clarity, improving cross-modal matching between events and LiDAR data.

\item We design a module that predicts an auxiliary variable as a regularization term during training, mitigating the negative effects of bias in the ground truth poses.


\end{itemize}

\section{Related Work}
In this section, we review the event-based visual localization techniques, including event-based visual place recognition and event-based camera pose relocalization.

\subsection{Event-based Visual Place Recognition}
Visual Place Recognition (VPR) is a subfield of visual localization, which aims to determine whether the current visual scene matches any previously visited. Typically, a set of geo-stamped reference images is given, and VPR techniques match the query images with them to identify the camera's location. A major challenge is that the same location can look drastically different due to changes in time, weather, season, or viewpoint. Event cameras, with their high dynamic range, are particularly suited to handle these variations, offering resilience to varying lighting conditions.

Milford et al. \cite{milford2015towards} first introduce event cameras for visual place recognition. Events are first accumulated into 10 ms time windows and then downsampled to low-resolution event frames. Finally, they use the common-used SeqSLAM algorithm\cite{milford2012seqslam} to perform place recognition. Fischer et al.\cite{fischer2020event} reconstruct multiple intensity image sets from event streams with different time spans or window lengths and propose an ensemble method to combine them for improved place matching. Kim et al.\cite{9635907} design a network to reconstruct denoised event edges from the event stream, and then introduce the classic Vector of Locally Aggregated Descriptors (VLAD) for place recognition. Kong et al.\cite{kong2022event} further extend this scheme to propose the first end-to-end event-based VPR network. Additionally, Fischer et al.\cite{fischer2022many} propose sampling only a limited number of pixels to construct descriptors, enabling computationally efficient VPR.

Event-based VPR typically requires a pre-established database consisting of images or events, which takes a lot of labor. In contrast, we utilize offline LiDAR maps as reference maps, which can be easily constructed using widely available consumer-grade LiDAR sensors. Furthermore, VPR provides only a coarse estimate of the camera's position, so it can be seen as the initial step of our method.

\subsection{Event-based Camera Pose Relocalization}
Camera Pose Relocalization (CPR) focuses on training a neural network tailored to a specific scene, enabling it to precisely estimate 6-DoF camera poses within the same scene used for training.
Event-based CPR methods use events as input and typically design their networks in a manner similar to conventional camera-based CPR approaches. Nguyen et al.\cite{nguyen2019real} first introduce event cameras for CPR. They first transform events within a short time interval into event frames and design a stacked spatial LSTM network to learn the camera pose. Jin et al.\cite{jin20216} incorporate a denoising model to reduce excessive noise in complex scenes, improving pose estimation accuracy. Similarly, Lin et al.\cite{lin20226} propose a Reversed Window Entropy Image (RWEI) generation framework to generate event frames with clear edges and utilize an attention-based DSAC* pipeline\cite{Brachmann2021VisualCR} to estimate camera poses. More recently, Ren et al.\cite{ren2024simple} present a method that directly utilizes the raw point cloud as network input, leveraging the high temporal resolution and inherent sparsity of events for more efficient pose estimation.

CPR techniques are popular for their high computational efficiency, making them well-suited for leveraging the low latency of event cameras. However, both conventional camera-based and event camera-based CPR methods implicitly rely on model parameters to encode the scene, which limits their scalability and generalization to new environments. In contrast, our method explicitly utilizes existing LiDAR maps as reference maps, allowing for more scalable deployment.

\vspace{2mm}
The most closely related work to ours is the work by Yuan and Ramalingam\cite{yuan2016fast}. They detect edges from events and represent the 3D model of the environment using vertical lines. Camera poses are then estimated by establishing correspondences between 2D event lines and 3D world lines. However, their approach only estimates 3-DoF poses and relies on the presence of abundant lines in the scene.

\section{PROPOSED METHOD}
In this section, we will present the details of our method. An overview of the proposed method is presented in Fig. \ref{fig:framework}.
\subsection{Event Representation}
\begin{figure}[h]
    \centering
    \includegraphics[width=0.95\linewidth]{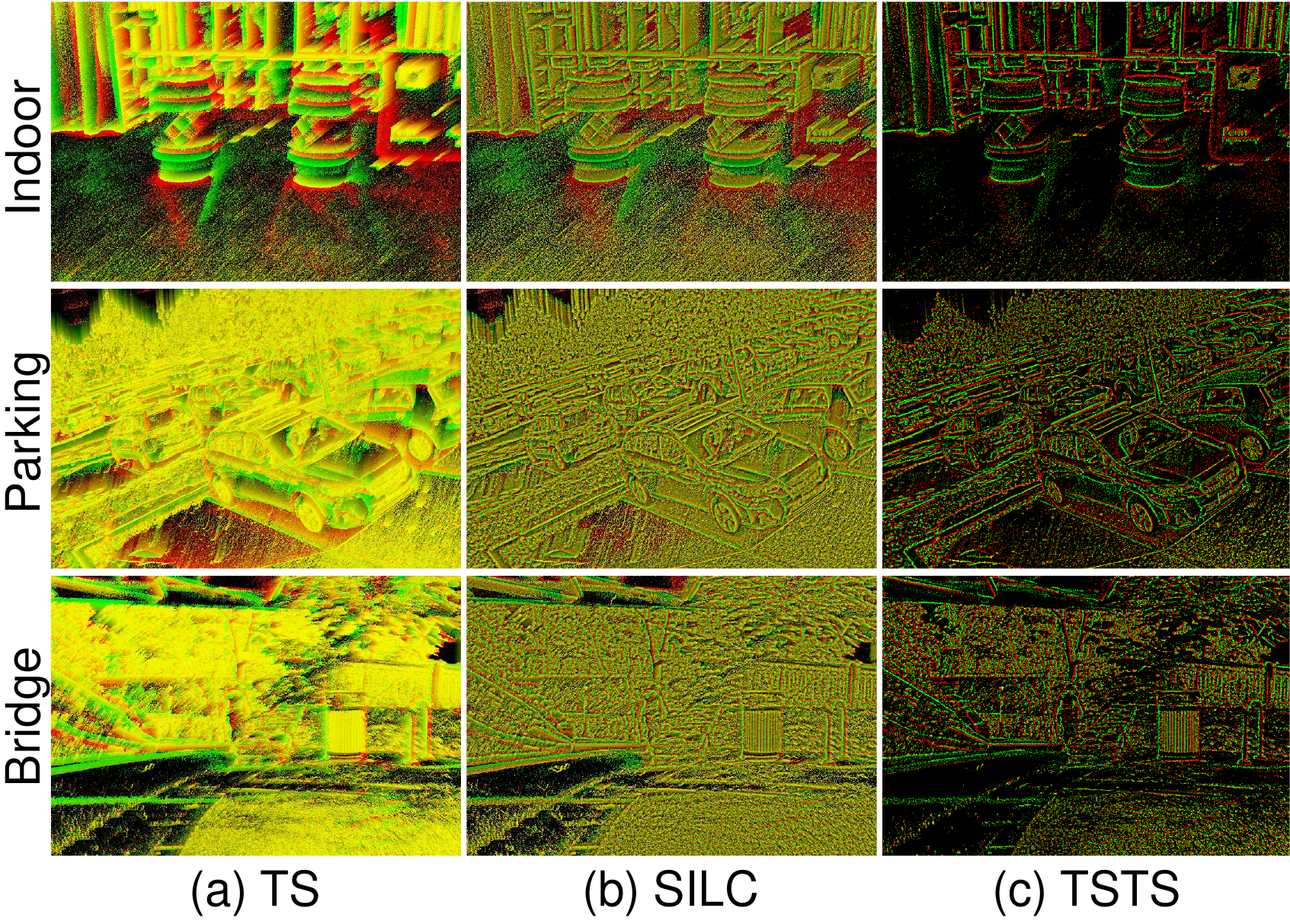}
    \caption{Comparison of resulting event frames from TS\cite{zhu2019unsupervised}, SILC\cite{manderscheid2019speed}, and our proposed TSTS.}
    \label{fig:event_frame}
    \vspace{-5mm}
\end{figure}

An event captured by an event camera is typically represented by a tuple $e_t=(x_t,y_t,p_t,t)$, where $(x_t,y_t)$ denotes its pixel location, $t$ denotes its timestamp, and $p_t$ denotes its polarity. To utilize existing neural networks to process the asynchronous events, a commonly used representation is Surface of Active Events\cite{benosman2012asynchronous}, also referred to as Time Surface\cite{lagorce2016hots}, defined as:
\begin{equation}
    T(x,y,p) \leftarrow t
\end{equation}
where $t$ is the timestamp of the latest event with polarity $p$ that occurred at pixel $(x,y)$. 
Typically, a fixed window length of events is accumulated to generate the time surface. This window length can be based on either a fixed time span or a fixed number of events. However, faster motion and richer textures both generate more events, making it challenging to choose an appropriate window length. A large window length causes blurring at the edges of the generated event frames, as shown in Fig. \ref{fig:event_frame} (a), while a small window length results in insufficient details. Existing methods often determine the window length based on a specific dataset\cite{sekikawa2019eventnet} or performance on a validation set\cite{maqueda2018event}, which is not trivial for practical applications. Fischer et al.\cite{fischer2020event} propose an ensemble scheme that combines multiple window lengths, but this approach adds more computational overhead. 


In this paper, we develop a novel event representation derived from the time surface, named Temporal-Spatial stable Time Surface (TSTS). Our method consists of two parts: deblur in the temporal dimension and denoise in the spatial dimension. Specifically, for each incoming event $e_t(x_t, y_t, p_t, t)$, we update the corresponding pixel $(x_t, y_t)$ in channel $p_t$ of the event frame $S(x,y,p)$ with the timestamp $t$. At the same time, we consider all valid pixels (non-zero) in the local neighborhood of  window size $2R$+$1$ around pixel $(x_t, y_t, p_t)$, defined as:
\vspace{-1mm}
\begin{equation}
    \mathcal{N}(x,y)_R \! = \! \{(x\!+\!dx,y\!+\!dy)|dx,dy\!\in \! \mathbb{Z}, |dx|,|dy|\! \leq \!R\}
\end{equation}
Each pixel in this neighborhood is adjusted by subtracting a value based on its difference from the central pixel's value. A parameter $\alpha$ is introduced to adjust the scale of this value. This process consistently suppresses older events while retaining the newer ones. Additionally, event cameras often capture excessive noise due to hardware limitations, which also impacts the structural clarity of the generated event frames. To address this, we design a denoising scheme based on an observation that meaningful areas in event frames tend to be denser. Specifically, for each incoming event $e_t(x_t, y_t, p_t, t)$, we count the number of valid pixels within a local neighborhood $\mathcal{N}(x_t,y_t)_r$ of window size $2r$+$1$. If this count is below a pre-defined threshold $\beta$, this event point is removed from the generated event frame. The final generated event frames are shown in Fig. \ref{fig:event_frame} (c). In our experiments, we sample 100 ms of events to generate each event frame and empirically set the parameters as follows: $R=6$, $\alpha=15$, $r=1$, and $\beta=0.7$ for all samples. The pseudo-code for the algorithm is provided in Algorithm \ref{alg}.


\vspace{-3mm}

\subsection{Depth Map Generation}
Given the initial pose $\boldsymbol{T}_{\text{init}}=(\boldsymbol{R}_{\text{init}}, \boldsymbol{t}_{\text{init}})$, the LiDAR points are transformed from the world coordinate system into the camera coordinate system using the transformation $P = \boldsymbol{T}_{\text{init}}^{-1} P_w$, where $\boldsymbol{T}_{\text{init}}^{-1}$ is the inverse of the initial camera pose, represented as:
\begin{equation}
    \boldsymbol{T}_{\text{init}}^{-1}=\begin{bmatrix}\boldsymbol{R}_{\text{init}}^{\top} & -\boldsymbol{R}_{\text{init}}^{\top}\boldsymbol{t}_{\text{init}} \\ 0 & 1 \end{bmatrix}
\end{equation} 
The next step is to project these transformed 3D points into 2D image space using the pinhole camera model:
\begin{equation}
    \left(\begin{array}{l}
    u \\
    v \\
    1
    \end{array}\right)=\frac{1}{Z}\left(\begin{array}{ccc}
    f_x & 0 & c_x \\
    0 & f_y & c_y \\
    0 & 0 & 1
    \end{array}\right)\left(\begin{array}{l}
    X \\
    Y \\
    Z
    \end{array}\right) \triangleq \frac{1}{Z} \boldsymbol{K} P,
\end{equation}
where $P=(X\;Y\;Z)^{T}$  represents the 3D coordinates in the camera coordinate system, and $(u\;v\;1)^{T}$ represents the corresponding homogeneous coordinates in the 2D image plane. The matrix $\boldsymbol{K}$ contains the camera's intrinsic parameters: $f_x, f_y, c_x$, and $c_y$. The depth value $I(u,v)$ is obtained by normalizing the $Z$ coordinate of $P$. To refine the depth map, we also apply the occlusion removal method outlined in \cite{Pintus2011RealtimeRO}, which eliminates occluded points from the final projection.
\begin{algorithm}[htbp]
\caption{Temporal-Spatial stable Time Surface}
\label{alg}
Output: Event Frame $S(x, y, p)$\\
Initialization: $S(x, y, p) \leftarrow 0$ for all $(x, y, p)$, $mask(x, y) \leftarrow 1$ for all $(x,y)$\\
For each incoming event $e(x, y, p, t)$, update $S$:
\begin{algorithmic}
\For{$-R \leq dx \leq R$}
\For{$-R \leq dy \leq R$}
\If{$S(x+dx, y+dy, p) \textgreater 0$}
\State $S(x+dx, y+dy, p) \leftarrow S(x+dx, y+dy, p) - (S(x,y,p) - S(x+dx, y+dy, p))/\alpha$
\EndIf
\EndFor
\EndFor
\State{$S(x, y, p) \leftarrow t$}
\State{Initialize $valid\_point=0$}
\For{$-r \leq dx \leq r$}
\For{$-r \leq dy \leq r$}
\If{$S(x+dx, y+dy, p) \textgreater 0$}
\State{$valid\_point = valid\_point + 1$}
\EndIf
\EndFor
\EndFor
\State{$valid\_rate = valid\_point / (2 \times r+1)^2$}
\If{$valid\_rate \textless \beta$}
\State{$mask(x,y)=0$}
\EndIf
\end{algorithmic}
$S(x,y,p) = S(x,y,p) \times mask$
\end{algorithm}
\vspace{-5mm}

\subsection{Flow Estimation Network}
After obtaining the event frame and the depth map, we input them into a neural network to estimate the event-depth flow. This network is based on the classic optical flow estimation model RAFT\cite{Teed2020RAFTRA}, with minor modifications\cite{CHEN2022209}. We replace the single feature encoder with two distinct encoders to handle the event frames and depth maps separately. Additionally, the input layer's channel configuration is adjusted to accommodate different inputs.

The ground truth event-depth flow for supervision is generated by calculating the distance between the depth map generated based on the ground truth pose and the initial pose respectively, denoted as follows:
\begin{equation}
    {\boldsymbol{f}_{\text{gt}}=\pi\left(P_w, {\boldsymbol{T}}_{\text {gt}}\right)-\pi\left(P_w, \boldsymbol{T}_{\text {init}}\right)},
\end{equation}
\begin{equation}
    \pi(P, \boldsymbol{T}) \triangleq \boldsymbol{K T} P,
    \label{formula:projection}
\end{equation}
where $\pi$ denotes the camera projection function that transforms 3D points into 2D space, while $\boldsymbol{f}_{\text{gt}}$ represents the generated ground truth event-depth flow. $\boldsymbol{T}_{\text{gt}}$ refers to the ground truth pose provided by the dataset. The loss function used for network supervision is defined as follows:
\begin{equation}
    \mathcal{L}=\frac{\sum(m(u, v)\left\|\boldsymbol{f}_{\text{pre}}(u, v)-\boldsymbol{f}_{\text{gt}}(u, v)\right\|_2)}{\sum m(u,v)},
\end{equation}
\begin{equation}
    m(u, v)=\left\{\begin{array}{l}1, \boldsymbol{f}_{\text{gt}} \neq 0 \\
            0, \text {otherwise}
\end{array}\right.,
\end{equation}
where $\boldsymbol{f}_{\text{pre}}$ denotes the predicted event-depth flow, while $\boldsymbol{f}_{\text{gt}}$ refers to the ground truth event-depth flow. $m(u,v)$ serves as a mask, identifying the valid pixels within the ground truth event-depth flow.

\vspace{-3mm}
\begin{figure}[h]
    \centering 
    \includegraphics[width=0.75\linewidth]{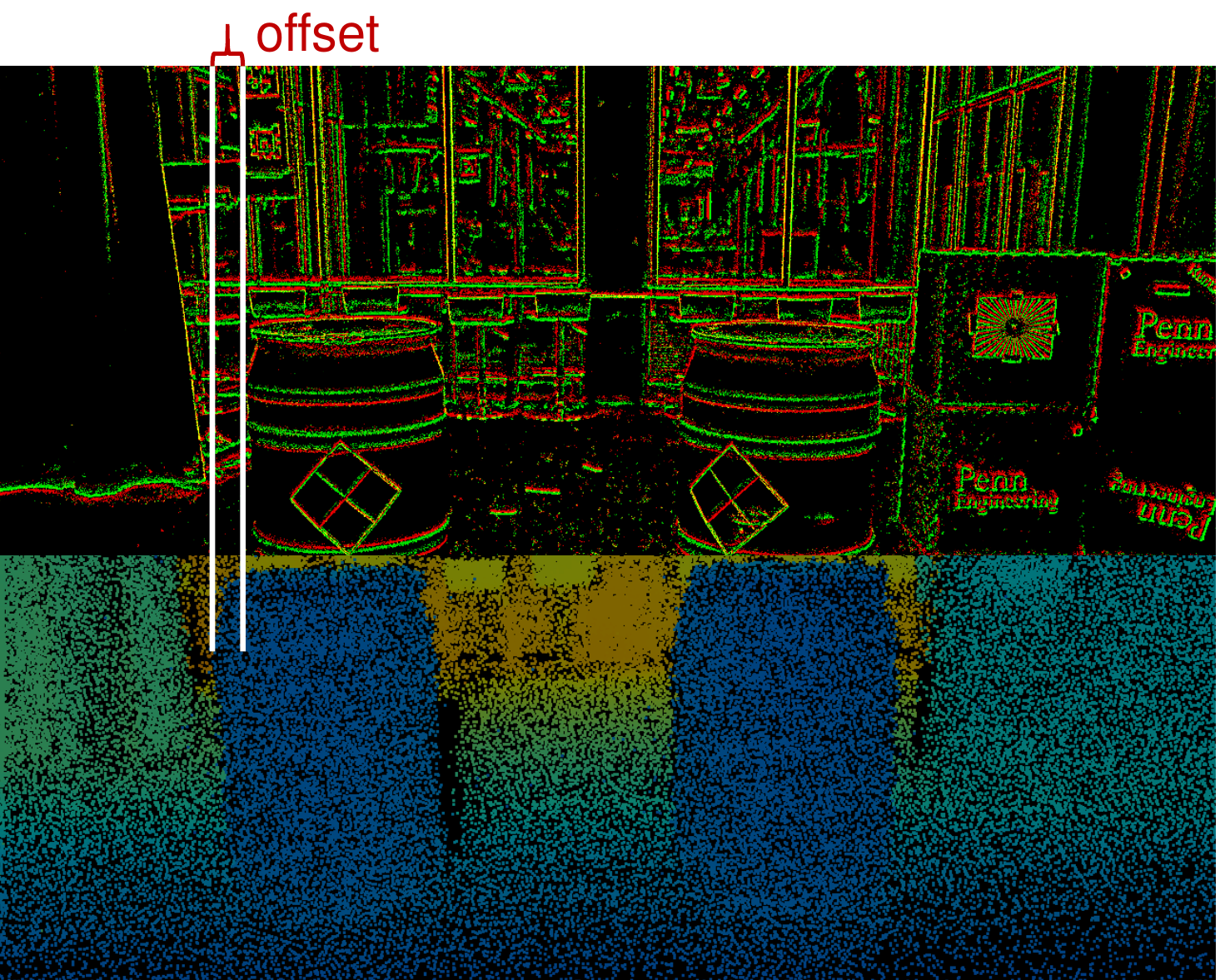}
    \caption{Offset exists between the event frame and the corresponding depth map generated based on the ground truth pose (in M3ED\cite{Chaney_2023_CVPR}).}
    \label{fig:offset}
    \vspace{-5mm}
\end{figure}

\subsection{Offset Alleviation Module}
\begin{figure}[h]
    \centering
    \includegraphics[width=0.8\linewidth]{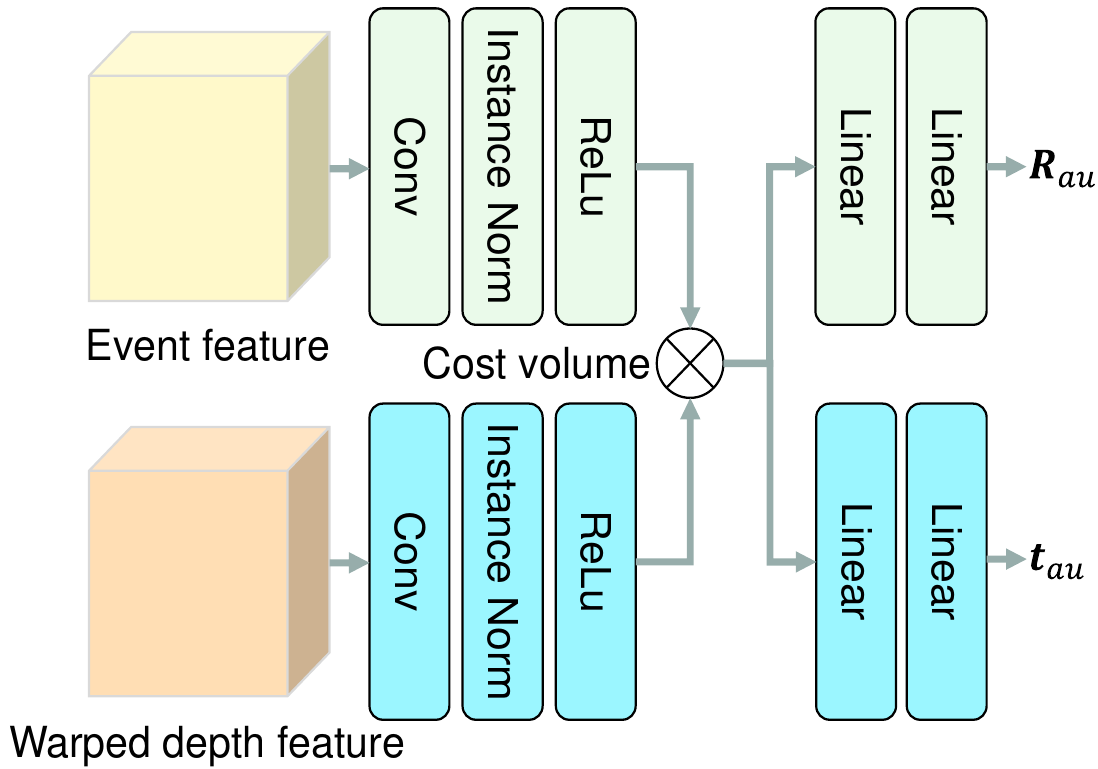}
    \caption{Overview of the devised offset alleviation module.}
    \label{fig:offset_estimator}
    \vspace{-7mm}
\end{figure}

Another critical issue is the offset between the event frame and the corresponding depth map generated using the dataset's ground truth pose (as shown in Fig. \ref{fig:offset}). This offset leads to imprecise event-depth flows during supervision, preventing the network from converging to a desired minimal.
A promising method to address this issue is adding an auxiliary variable for each training example\cite{hu2019simple}. We follow this idea to introduce a simple module. Specifically, after estimating the event-depth flow for each sample, we use it to warp the depth feature map and then compute the cost volumes between the warped depth feature map and the event feature map. These cost volumes are then passed through two fully connected layers to estimate the 6-DoF auxiliary variable $\boldsymbol{T}_{\text{au}}=(\boldsymbol{R}_{\text{au}}, \boldsymbol{t}_{\text{au}})$. 
The network structure of the estimator is shown in Fig. \ref{fig:offset_estimator}.
This variable is added to the ground truth pose to compute the event-depth flow for supervision, denoted as follows:
\begin{equation}
    {\boldsymbol{f}_{\text {gt}}^{*}=\pi\left(P_w, {\boldsymbol{T}}_{\text {gt}}\boldsymbol{T}_{\text{au}}\right)-\pi\left(P_w, \boldsymbol{T}_{\text {init}}\right)},
\end{equation}
\begin{equation}
    \mathcal{L}=\frac{\sum (m(u, v)\left\|\boldsymbol{f}_{\text{pre}}(u, v)-\boldsymbol{f}_{\text{gt}}^{*}(u, v)\right\|_2)}{\sum m(u,v)},
\end{equation}
where $\boldsymbol{f}_{\text {gt}}^{*}$ represents the ground truth event-depth flow compensated by the estimated auxiliary variable. This variable helps alleviate the offset issue, enhancing the network’s ability to learn event-depth matching.


Finally, we calculate the camera pose $\boldsymbol{T}_{\text{pre}}$ using a PnP solver based on the estimated event-depth flow, i.e., 2D-3D correspondences. For this, we utilize the publicly available PoseLib library\cite{PoseLib}. The maximal reprojection error is set to 12.0, and we apply the Huber function for optimization, while other parameters are left at their default values. 

\section{Experiments}
In this section, we introduce the datasets, the evaluation metrics, the experimental setup, and the experimental results.

\subsection{Datasets and Evaluation Metrics}
\subsubsection{Datasets}
We conduct experiments on M3ED\cite{Chaney_2023_CVPR} and MVSEC\cite{zhu2018multivehicle}. The M3ED dataset includes three platforms: Falcon, Spot, and Car, each used to capture data across various indoor and outdoor scenes. A key criterion for selecting sequences in our experiments is the quality of the LiDAR maps. LiDAR points can shift in scans due to velocity uncertainties of moving objects, introducing excessive noise into the constructed maps. Therefore, we select specific sequences in our experiments. We primarily use \textit{falcon\_indoor\_flight} 1 and 2 for network training, and \textit{falcon\_indoor\_flight} 3 for testing. Additionally, we fine-tune the trained model on \textit{falcon\_outdoor\_day\_penno\_parking} 1 and \textit{spot\_outdoor\_day\_srt\_under\_bridge} 1 respectively, then correspondingly test it on \textit{falcon\_outdoor\_day\_penno\_parking} 2 and \textit{spot\_outdoor\_day\_srt\_under\_bridge} 2 to evaluate the performance of our method in outdoor scenes. The MVSEC dataset contains event data in flying, driving, and handheld scenes. The event cameras in MVSEC have a resolution of $346 \times 260$, which is relatively low compared to the $1280 \times 720$ resolution in M3ED. For evaluation, we retrain our model on \textit{indoor\_flying} 1, and then test it on \textit{indoor\_flying} 2 and 3. 

\subsubsection{Evaluation Metrics}
For evaluation, we create a set of initial pose guesses by adding deviation to the ground truth by [-50 cm, +50 cm] in translation and [-5°, +5°] in rotation, simulating the uncertainty of a preceding coarse localization step. In our experiments, we independently introduce random disturbances to each ground truth pose to generate the initial estimates. Localization accuracy is evaluated using the mean and median errors of the predicted poses.

\subsection{Experimental Setup}
For the sake of reducing computational overhead, we limit all frames (including generated event frames and depth maps) to a resolution of $960 \times 600$. As mentioned above, we utilize modified RAFT as the backbone. We first train the model on falcon\_indoor\_flight 1 and 2 for 100 epochs with $batch\_size=2, weight\_decay=1e^{-4}, learning\_rate=4e^{-5}$, and $learning\_rate\_scheduler=\emph{OneCycleLR}$. For fine-tuning on other sequences in M3ED, we conduct training on the corresponding data for an additional 30 epochs with constant $learning\_rate=4e^{-5}$. For other datasets, we retrain the model for 100 epochs following the same setup as used for M3ED. A single NVIDIA GTX 4090 GPU is used for all experiments.

\begin{table}[h]
    \centering

    \caption{Ablation experiments. We only adjust the input layer's channel configuration to accommodate different inputs. All models are tested on sequence \textit{falcon\_indoor\_flight} 3 in M3ED.}
    \label{tab:1}
    \resizebox{1.0\linewidth}{!}{
    \begin{tabular}[t]{c|c|cc|cc}
        \toprule
        \toprule
        \multirow{2}{*}{Case}&\multirow{2}{*}{Representation} & \multicolumn{2}{c|}{Mean} & \multicolumn{2}{c}{Median}\\
        & & Transl.[cm]$\downarrow$ & Rot.[°]$\downarrow$ &Transl.[cm]$\downarrow$ & Rot.[°]$\downarrow$\\ 
        \midrule
        (a) & Grayscale & 18.87 & 2.76 &17.35 & 2.37\\
        \midrule
        (b) & Voxel Grid \cite{zhu2019unsupervised} & 10.52 & 1.32 &8.79 & 1.03\\
        \midrule
        (c) & TS\cite{benosman2012asynchronous} & 10.04 & 1.22 &8.57 & \textbf{0.89}\\
        \midrule
        (d) & SILC\cite{manderscheid2019speed} & 12.74 & 1.58 & 10.35 & 1.19\\
        \midrule
        (e) & Ours (TSTS) & 9.77 & 1.20 & 8.05 & \textbf{0.89}\\
        \midrule
        (f) & Ours (TSTS) + OAM & \textbf{9.34} & \textbf{1.17} & \textbf{7.78} & \textbf{0.89}\\
        \bottomrule
        \bottomrule
    \end{tabular}
    }
    \vspace{-5mm}
\end{table}

\subsection{Ablation Study}
We begin by conducting ablation experiments to assess the impact of the proposed modules. The results are shown in Table \ref{tab:1}. The model is trained on data from both the left and right cameras in the \textit{falcon\_indoor\_flight} 1 and 2 sequences, and then tested using event data from the left camera in \textit{falcon\_indoor\_flight} 3.

\begin{table*}[!htbp]
    \caption{Comparison with conventional camera-based visual localization method I2D-Loc\cite{CHEN2022209} in various scenes from several public datasets. For the sake of saving space, we abbreviate the sequences 
    \textit{falcon\_outdoor\_day\_penno\_parking} 2 and \textit{spot\_outdoor\_day\_srt\_under\_bridge} 2 in M3ED to \textit{falcon\_o\_d\_p\_p\_2} and \textit{spot\_o\_d\_s\_u\_b\_2}, respectively.
    }
    \centering
    \resizebox{1.0\linewidth}{!}{
    \begin{tabular}{c|c|cc|cc|cc|cc}
        \toprule
        \toprule
        & & \multicolumn{4}{c|}{I2D-Loc} & \multicolumn{4}{c}{Ours}\\
        \multirow{2}{*}{Dataset} & \multirow{2}{*}{Sequence} & \multicolumn{2}{c|}{Mean} & \multicolumn{2}{c|}{Median} & \multicolumn{2}{c|}{Mean} & \multicolumn{2}{c}{Median} \\
        & & Transl.[cm]$\downarrow$ & Rot.[°]$\downarrow$ &Transl.[cm]$\downarrow$ & Rot.[°]$\downarrow$ & Transl.[cm]$\downarrow$ & Rot.[°]$\downarrow$ &Transl.[cm]$\downarrow$ & Rot.[°]$\downarrow$\\
        \midrule
        \multirow{3}{*}{\textbf{M3ED}\cite{Chaney_2023_CVPR}}
        & \textit{falcon\_indoor\_flight} 3 &18.87 &2.76& 17.35& 2.37& \bf{9.34} & \bf{1.17} & \bf{7.78} & \bf{0.89}\\
        & \textit{falcon\_o\_d\_p\_p\_2} & 39.79& 3.67& 39.39& 3.31& \bf{33.67} & \bf{1.95} & \bf{30.77} & \bf{1.50}\\
        & \textit{spot\_o\_d\_s\_u\_b\_2} & 35.54& 3.02& 33.41& 2.75& \bf{27.12} & \bf{1.79} & \bf{25.15} & \bf{1.53} \\
        \midrule
        \multirow{2}{*}{\textbf{MVSEC}\cite{zhu2018multivehicle}}&\textit{indoor\_flying} 2 & 29.07& 3.99& \bf{24.16} & 3.56& \bf{26.99} & \bf{3.61} & 24.64 & \bf{3.39}\\  
        & \textit{indoor\_flying} 3 & 18.58 & 2.46 & 16.70 & 2.26 & \bf{13.90} & \bf{2.09} & \bf{12.03} & \bf{2.00}
        \\
        \toprule
        \toprule
    \end{tabular}
    }
    \label{tab:2}
    \vspace{-7mm}
\end{table*}

To validate the effectiveness of the proposed event representation TSTS, we compare the performance of EVLoc using different input representations, as shown in Table \ref{tab:1} (a)-(e). Initially, we train the model using grayscale images as input. When we replace this with the Voxel Grid\cite{zhu2019unsupervised}, a commonly used event representation, both the translation and rotation errors significantly decrease. This result confirms the superiority of event cameras for localization in challenging scenes. Next, we replace the Voxel Grid representation with the classic Time Surface\cite{benosman2012asynchronous}. The results indicate an improvement in localization accuracy, suggesting that the Time Surface is better suited for our framework. We attribute this to the Time Surface’s emphasis on the latest events, which helps the structures in the generated event frames align more closely with the real-world scene at the given timestamp. 
However, due to the complexity of motion, it is challenging to determine an optimal window length for all samples, which often results in motion blur in the generated event frames. Then we try using the Speed-Invariant Time Surface (SILC)\cite{manderscheid2019speed} as input, but the result is worse than that with the original Time Surface. We believe this is because the SILC representation is designed to be invariant to an object's local speed, but it does not prioritize preserving clear edges. Finally, we use our proposed event representation TSTS as input, and the experimental results show a significant improvement in localization performance, confirming the effectiveness of our approach.

We further conduct experiments to validate the effectiveness of the proposed offset alleviation module. The results are shown in Table \ref{tab:1} (f).  We incorporate the offset alleviation module during both training and inference for our event representation TSTS. The results demonstrate that the module enhances the network's performance, confirming the effectiveness of the proposed offset alleviation module.

\subsection{Results Analysis}
\begin{figure}[htbp]
	\centering
	\begin{minipage}[b]{0.49\linewidth}
		\centering
		\includegraphics[width = 1.0\columnwidth]{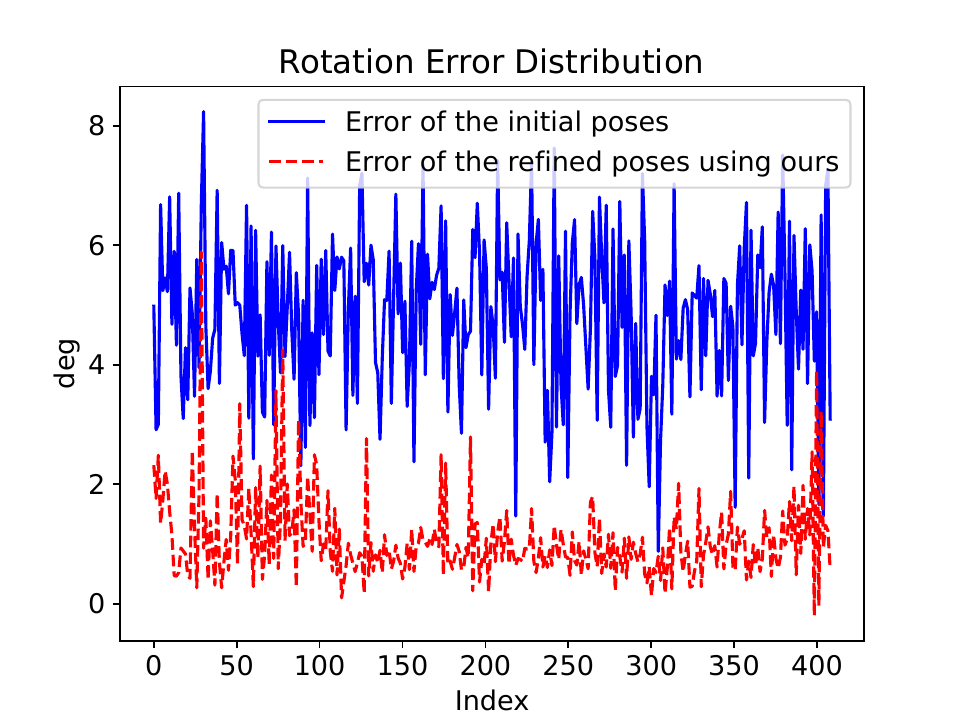} 
		\subcaption{Rotation}
	\end{minipage}
	\begin{minipage}[b]{0.49\linewidth}
		\centering
		\includegraphics[width = 1.0\columnwidth]{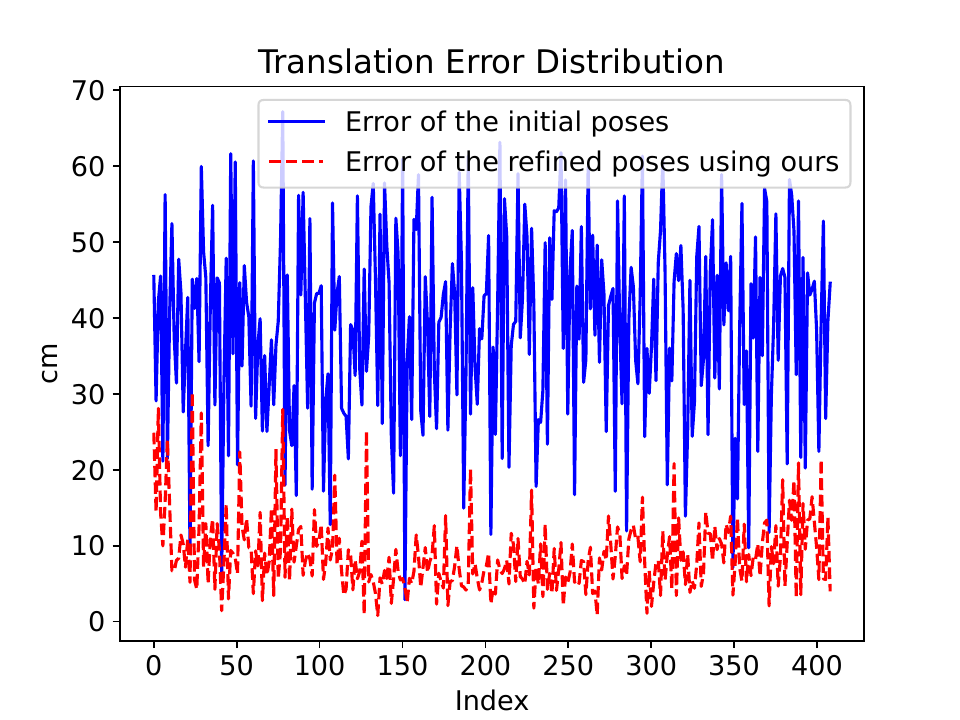} 
		\subcaption{Translation}
	\end{minipage}
    \caption{Diagram of error distribution. The results come from the experiments on M3ED \textit{falcon\_indoor\_flight} 3. The blue line indicates the errors of the initial poses, while the red line shows the errors of the refined poses after using our proposed method.}
    \label{fig:distribution}
    \vspace{-6mm}
\end{figure}

As shown in Fig.\ref{fig:distribution}, we visualize the error distribution of the initial and estimated poses using our method. The comparison demonstrates that our approach significantly reduces pose errors in both rotation and translation for the majority of samples. The improvement is not significant for some samples since there exists little overlap between the depth map and the event frame. 

Furthermore, we compare our method with the advanced conventional camera-based method I2D-Loc\cite{CHEN2022209} on various scenes from multiple datasets to assess the effectiveness of our method in more complex environments.
The experimental results are shown in Table \ref{tab:2}. We first conduct experiments on other sequences within the M3ED dataset. As outlined in the experimental setup, we select scenes with multiple sequences, fine-tune the model on one sequence, and test it on the others. These selected scenes include \textit{falcon\_outdoor\_day\_penno\_parking} and \textit{spot\_outdoor\_day\_srt\_under\_bridge}, both set in outdoor environments with complex lighting conditions. 
The results show that our method outperforms I2D-Loc in these outdoor scenes. This is because some contours in grayscale images can be lost in overexposed areas or deep shadows due to the limited dynamic range of conventional cameras, while they remain visible in event frames. We can also see that the performance of both methods is slightly lower compared to indoor scenes, likely due to increased noise and the less structured nature of outdoor environments, making feature matching more challenging.

Additionally, we train our model and the baseline on the MVSEC dataset as well, which is commonly used for event-based experiments. Though similar outperformance over baseline can be observed, both methods demonstrate performance drops in this dataset.
This is due to the limited quality of the LiDAR data in the dataset. There exists a lot of holes in the LiDAR maps, making the generated depth maps lack sufficient information for accurate matching. Despite the limited localization performance in these more challenging scenarios, the results confirm the effectiveness of our proposed method. They also highlight that the quality of LiDAR maps is just as crucial as the event data for event-based visual localization in LiDAR maps.

\vspace{-3mm}

\section{Conclusions and Future Work}
We propose a novel framework for event-based visual localization using LiDAR maps. The core idea is to leverage an optical flow estimation network to align events with LiDAR points in 2D space. We also introduce a new frame-based event representation that enhances structural clarity, improving the matching process between the two modalities. Additionally, we address offset issues caused by imprecise ground truth poses with a dedicated regularization scheme. Experimental results across multiple scenes from two public datasets validate the effectiveness of our approach. 
\textbf{Limitations:} Several open challenges remain as potential directions for future work. 
First, our approach still relies on an initial pose for localization. In the future, we may integrate global localization into our method, enabling fully end-to-end localization.
Second, the current framework relies on the existence of overlap between events and LiDAR points in 2D space.
Third, instead of solely improving event processing methods, a promising avenue is to enhance the extraction of useful information from LiDAR maps, as they are often imperfect.

\section{Acknowledgement}
This research project has been supported by the Austrian Science Fund (FWF) under project agreement (I 6747-N) EVELOC.

\newpage
\balance
\bibliographystyle{IEEEtran.bst}
\bibliography{references}
\end{document}